\pdfoutput=1
\documentclass[11pt]{article}

\usepackage{ACL2023}

\usepackage{times}
\usepackage{latexsym}
\usepackage{enumitem}
\usepackage[T1]{fontenc}
\usepackage[utf8]{inputenc}
\usepackage{pifont}
\usepackage{adjustbox}
\usepackage{tcolorbox}
\usepackage{graphicx}
\graphicspath{ {./} }

\usepackage{microtype}

\usepackage{inconsolata}

\title{On the Challenges of Building Datasets for Hate Speech Detection}

\author{Vitthal Bhandari \\
  Birla Institute of Technology and Science, Pilani, India \\
  \texttt{f20170136p@alumni.bits-pilani.ac.in} \\}

\begin{document}
\maketitle
\begin{abstract}
Detection of hate speech has been formulated as a standalone application of NLP and different approaches have been adopted for identifying the target groups, obtaining raw data, defining the labeling process, choosing the detection algorithm, and evaluating the performance in the desired setting. However, unlike other downstream tasks, hate speech suffers from the lack of large-sized, carefully curated, generalizable datasets owing to the highly subjective nature of the task.

In this paper, we first analyze the issues surrounding hate speech detection through a data-centric lens. We then outline a holistic framework to encapsulate the data creation pipeline across seven broad dimensions by taking the specific example of hate speech towards sexual minorities. We posit that practitioners would benefit from following this framework as a form of best practice when creating hate speech datasets in the future.
\end{abstract}

\section{Introduction}

In the last couple of years, the NLP community has witnessed an increased interest in detecting hateful and toxic speech owing to the wide scope and nature of the task. A number of workshops (such as SemEval) have been dedicated to the promote the research in this domain \citep{kirk2023semeval, perez-almendros-etal-2022-semeval, fersini-etal-2022-semeval, pavlopoulos-etal-2021-semeval} and several shared tasks have been organized to help make sense of popular datasets \citep{magnossao-de-paula-etal-2022-upv, bhandari-goyal-2022-bitsa}.

However, hate speech detection is an inherently subjective task. This simply means that datasets and language models developed for a particular setting and trained to serve a specific objective do not generalize well to other settings, even if it appears that they do. This renders standalone speech tasks and datasets in this domain impractical for a wide variety of applications, unknown to the knowledge of the na\"ive user. To ensure that models behave robustly, reliably, and fairly when making predictions about data that is different from the data they were trained on, it is crucial to have specific knowledge of the choices made by the dataset creators along the entire pipeline.

Work has been done focusing on different parts of the pipeline which are indicative of the intricacies involved in each step. For example, a number of researchers have discussed how unintended annotator bias affects the overall polarity of the resultant dataset. This calls for dataset authors to take steps to avoid annotator bias and enforce uniform rules during the annotation process.

\begin{figure*}[htbp]
\begin{center}
\includegraphics[width=\textwidth]{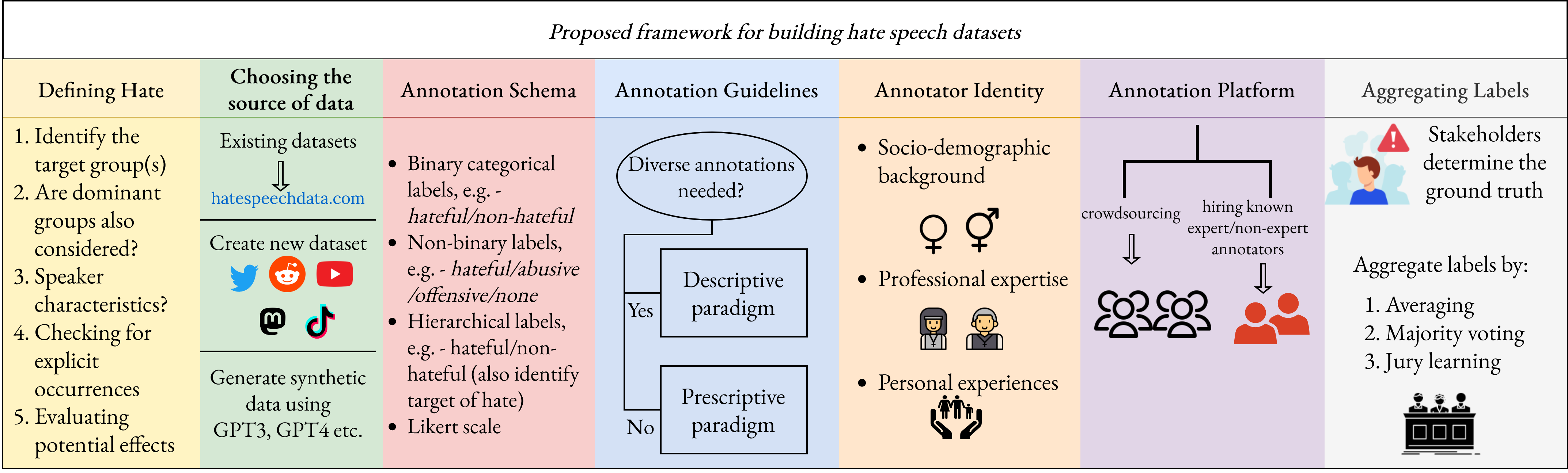}
\caption{Our proposed framework summarizing seven research agendas for building hate speech datasets}
\label{fig:fig1}
\end{center}
\end{figure*}

Our work attempts to weave together all such variables into a unified framework by presenting a data-driven taxonomy that highlights how all these concepts are linked and how they differ from one another. We start with a brief discussion of the issues commonly plaguing hate speech detection systems through the point of view of data (\S\ref{Related Work}). We then define the need for a consolidated framework that future dataset creators must follow (\S\ref{The Need for a Unified Framework}). Then we give a detailed description of our framework (distilled in Figure \ref{fig:fig1}) by providing a list of the different factors to be taken into consideration when building a dataset (\S\ref{Proposed Hate Speech Framework}). We finally discuss the open challenges in the domain of hate speech detection (\S\ref{Challenges}) before concluding our study (\S\ref{Conclusion}).

To summarize, our main contributions are as follows: (1) We look at hate speech detection through the lens of data and discern various problems that arise when datasets are not generalizable and data statements are missing; (2) We propose an overarching research agenda across seven different potential points of contention over the course of dataset creation; and (3) We summarize this research agenda as a form of best practices for researchers and practitioners to follow.

\section{Related Work}
\label{Related Work}

Prior research on identifying issues in hate speech detection has been broad and distinct. \citet{vidgen2020directions} conduct a data-driven analysis of 63 public hate speech datasets and provide recommendations to create new training datasets for abusive content. We build on their work and previous work on combining hate speech detection with human-centered design to build a comprehensive framework that takes all pain points into consideration when building a dataset.

The definition of \textit{hate} varies abundantly in the wild. \citet{fortuna-etal-2020-toxic} find that different definitions of hate are used for similar concepts such as abuse, aggression, and toxicity and this makes many publicly available datasets incompatible. \citet{yuan-etal-2022-separating} introduce an adversarial debiasing setup to help classifiers differentiate between hate speech and offensive language. \citet{khurana-etal-2022-hate} build on prior work to introduce criteria for defining \textit{hate} in order to clearly establish the purpose of annotations.

Even if the definition of hate is precise, the lack of proper context, as is the case with most datasets, decreases clarity for the annotators and introduces unintended bias \citep{mosca-etal-2021-understanding, yu-etal-2022-hate, sen-etal-2022-counterfactually}. Thus there is a need for strict guidelines to help annotators distinguish between hateful and non-hateful text in the absence of context. \citet{maronikolakis-etal-2022-analyzing} show that tweets containing African American English (AAE) and AAE+masculine content are disproportionately labeled as hateful more often. Well-formulated guidelines should help annotators avoid personal bias and allow for more consistent annotations. To this end \citet{rottger-etal-2022-two} define two different ways to frame annotation guidelines.

The annotation process itself involves several sensitive choices that affect the overall dataset. \citet{parmar-etal-2023-dont} find that recurring patterns in the crowdsourcing instruction examples cause crowd workers to propagate \textit{instruction bias} to the dataset wherein the model trained on that data fails to generalize beyond the said bias. \citet{sandri-etal-2023-dont} introduce a two-layered taxonomy for classifying annotators' disagreement in subjective tasks. They find that disagreements caused by personal bias and those caused by swearing have the highest prevalence in offensive language detection. 

A number of other review papers comprehensively discuss the available resources for and obstacles in hate speech detection and offer a reasonable starting point for research in this topic \citep{Poletto2020ResourcesAB, 10.1145/3232676, schmidt-wiegand-2017-survey, Kovcs2021ChallengesOH}. However, to the best of our knowledge, our work is the first of its kind that provides a thorough insight into challenges in hate speech detection \textit{w.r.t} the \textbf{data}. Most prior works focus on improving the metric-based performance of hate speech detection systems instead.

\section{The Need for a Unified Framework}
\label{The Need for a Unified Framework}

When building a dataset of hateful textual instances, the author is tasked with the burden of multiple choices at every single step of the process, starting from whether or not to label the data to writing down the annotation guidelines and aggregating the labels. Unfortunately, leaving many (or some) of these choices unaddressed, as has mostly been the case in NLP so far, leads to datasets containing undesirable characteristics such as annotator bias, weak inter-annotator agreement, a vague and unclear idea of the overarching theme, or even privacy concerns over the data.

These issues commonly plague NLP datasets in general. However, when dealing with subjective tasks such as hate speech detection we must not leave room for ambiguity during the process of data curation. We argue that \textbf{making informed choices} along each checkpoint in the process and \textbf{explicitly revealing them} as part of the data statement helps researchers evaluate in the future if the dataset can be used for other tasks or not. 

In the following section, we lay down a framework outlining an exhaustive research agenda that dataset creators must refer to in order to ensure that their datasets can be used fairly and reliably by others.

\section{Proposed Hate Speech Framework}
\label{Proposed Hate Speech Framework}

This section provides our proposed framework to reliably build hate speech datasets. As represented in Figure \ref{fig:fig1}, we define the following seven checkpoints, to completely encapsulate the scope of dataset creation when dealing with hateful text:

\begin{enumerate}[label={(\arabic*)}]
  \item Constructing the definition of \textit{hate}
  \item Choosing the right source of hateful text
  \item Labeling the data: defining the annotation schema
  \item Writing down the annotation guidelines
  \item Setting up the labeling process
  \item Sampling a suitable set of annotators
  \item Aggregating the labels
\end{enumerate}

For each agenda, we highlight the considerations that should be taken into account. 
It must be noted here that one need not iterate the framework linearly since a few agendas overlap with each other in terms of applicability. It might make sense to choose the data source first, write the annotation guidelines next, and define hate during this process.

To provide relevant context for the steps explained below, we will take the example of a sample task \textbf{T}: let us assume a group wants to analyze the spread of hate against sexual orientation minorities. They decide to build a dataset from a popular social media app and wish to annotate the dataset. They aim to have diverse opinions reflected in their dataset. As of now, they have no other design specifications in mind. Let us call this \textit{control dataset} \textbf{D}. As we move forward through the subsequent sections, we will use the proposed framework to give detailed examples of what choices can be made to design \textbf{D} along all possible steps. We urge the readers to follow a similar approach when designing datasets in the future.

\subsection{Defining Hate}
\label{Defining Hate}

Before building a dataset, i.e., before even collecting the data, it is important to define what \textit{hateful} means specific to the task at hand.

\citet{khurana-etal-2022-hate} break down \textit{hate} into smaller chunks (modules) and propose five criteria to be taken into account when \textit{constructing} a task-specific definition of hate: (1) target groups, (2) social status of target groups, (3) perpetrator, (4) type of negative reference, (5) type of potential consequence. They argue that by defining hate in a modular fashion, it is possible to vary the precision of the definition between vague and open to interpretation to clear and highly detailed.

To construct a task-specific definition of hate for \textbf{D}, the target group for the task would be \textit{sexual orientation}, and we could explicitly exclude the dominant groups from our consideration. The researchers may choose to not collect the speaker characteristics. We also require that the text contains at least one of stereotype, group characteristic, and a slur. Finally, the text must insult sexual minorities, or incite either violence, hate, or discrimination.

\begin{figure}[htbp]
    \hypertarget{RA1}{}
    \begin{tcolorbox}[colback=yellow!10!white, colframe=yellow!70!black, title=Research Agenda 1]
    Dataset creators must clearly define what is meant by \textit{hate} specific to the task to ensure that annotators will deviate from the given instructions only for those criteria. 
    \end{tcolorbox}
\end{figure}

\subsection{Choosing the Data Source}

When deciding what data to work with, it is best to first analyze existing datasets for similarities with the task at hand. For task \textbf{T} the researchers might go through the literature to scrounge for hate speech datasets which (1) have been extracted from a social media source (2) have the same target group (\textit{sexual minorities}), and (3) follow an annotation schema conducive to their research plan.

\texttt{\href{https://hatespeechdata.com/}{hatespeechdata.com}} is an open-source catalog of datasets annotated for hate speech and online abuse. The repository hosts more than 110 datasets across 25 different languages. It serves as a comprehensive starting point for anyone who wishes to explore the existing datasets.

However, sometimes the existing datasets do not capture the requirements or complexity of the task. In such a case it becomes difficult to use an off-the-shelf dataset to model a classifier. To take the example of task \textbf{T}, only a handful of datasets across the entire literature focus on sexual minorities as the target group. It then makes sense to build a custom dataset that satisfies the problem statement accurately.

Social media data is easily available for building modern datasets and provides flexibility in terms of querying, filtering, and analyzing data. It is also cheaper and more accessible compared to traditional forms of text such as news articles and journals. Twitter\footnote{\url{https://developer.twitter.com/en}}, Reddit\footnote{\url{https://www.reddit.com/dev/api/}}, YouTube\footnote{\url{https://developers.google.com/youtube/v3/docs}}, Mastodon\footnote{\url{https://docs.joinmastodon.org/api/}}, and TikTok\footnote{\url{https://developers.tiktok.com/}} expose web-client APIs for developers to query historical data of the order of billions of entities. These APIs are free of cost for most use cases but might impose rate limits on the number of hits and records fetched per unit time. Recent work has focused on certain alt-right extremist platforms such as Gab and Parler \citep{israeli-tsur-2022-free, mathew2021hatexplain, kennedy_atari_davani_yeh_omrani_kim_coombs_havaldar_portillo, qian-etal-2019-benchmark} since they allow greater freedom of speech and hence are rich reservoirs of hateful content. Most developer APIs allow keyword-based search across entire text or hashtags. Several third-party libraries and wrappers have also been built on top of these APIs and allow users to query with a finer level of granularity.

In certain situations, it becomes difficult to use either existing datasets as they are inadequate or create new datasets due to a lack of complete control over the querying process. For instance, to collect a set of hateful and sarcastic tweets, a simple keyword search is not enough. Each fetched tweet needs to be manually reviewed for the presence of sarcasm. This does not scale well to thousands of tweets. It's best to generate synthetic datasets in this case. \citet{hartvigsen-etal-2022-toxigen} introduce a dataset of 274k statements generated by GPT-3 \citep{brown2020language} using demonstration-based prompting and pretrained toxicity classifiers for \textit{\textbf{implicit}} hate speech detection.

\begin{figure}[htbp]
    \hypertarget{RA2}{}
    \begin{tcolorbox}[colback=green!8!white, colframe=green!55!black, title=Research Agenda 2]
    Dataset creators could either use existing datasets or create new ones by mining social media data or generate synthetic examples using generative models depending on the availability and novelty of data.
    \end{tcolorbox}
\end{figure}

\subsection{Defining the Annotation Schema} \label{Defining the annotation schema}

Annotation schema refers to the choice and structure of labels for a particular labeling task. It defines the range of discrete or continuous sets of values that each text label can assume. 

Detection of hate speech has historically been modeled as a binary classification task. This reduces the annotation schema to two values of \textit{hateful/non-hateful} or \textit{toxic/non-toxic} \citep{gao-huang-2017-detecting, Ribeiro_Calais_Santos_Almeida_Meira, Sarkar_KhudaBukhsh_2021}. To allow more flexibility for annotators, researchers have also used non-exclusive, non-binary ordinal or categorical labels. \citet{Founta_Djouvas_Chatzakou_Leontiadis_Blackburn_Stringhini_Vakali_Sirivianos_Kourtellis_2018} labeled abusive behavior on Twitter using four labels - Abusive, Hateful, Normal, Spam (categorical scale) whereas \citep{kennedy2020constructing} classified hate using ordinal ratings over ten different categorical components of hate. 

In reality, a surface-level annotation schema poses challenges in capturing the overall complexity and meaning of hate. It makes sense to devise a hierarchical taxonomy of hate instead of adhering to broad labels. To this end, many dataset creators have adopted a multi-level approach toward defining a more nuanced typology of hate speech. \citet{vidgen-etal-2021-introducing} present a taxonomy of six granular categories to differentiate between abusive (Identity-directed abuse, Affiliation-directed abuse, Person-directed abuse) and non-abusive (Non-hateful Slurs, Counter speech, Neutral) content. They further provide sub-categories into these categories.

Recent work highlighting the subjectivity of hate speech as an NLP task has motivated researchers to model hate as a range of continuous values across a scale. This type of schema: (1) allows for  more fine-grained labeling of the dataset, (2) increases expressiveness, and (3) can handle ambiguous or borderline content by allowing annotators to express their uncertainty by assigning values corresponding to their level of confidence instead of making a strict categorization. The schema could be a Likert scale from 1-3, or 1-5. \citet{cercas-curry-etal-2021-convabuse} rate abusive content on a scale of -3 to 1 depending on the severity of abuse with -3 being most abusive. However, another popular commercial classifier in the form of Google's Perspective API\footnote{\url{https://www.perspectiveapi.com/}} uses a normalized score in the range [0, 1] to label any text across different dimensions of hate. During training, the API uses the probability of the text containing a particular attribute as the ground truth instead of the mean or majority of all annotations for that text, i.e., if 3 out of the 10 annotators for a particular text feel it is toxic, the text will be assigned a score of 0.3 for the \texttt{TOXICITY} attribute.

In the scenario of task \textbf{T} it could be that the purpose of building \textbf{D} is: (1) to create a novel dataset that captures the complexity of hate, and (2) to encode diversity in the annotations - hence it might be difficult to achieve majority consensus on ground truth labels. Owing to these two requirements, we suggest that the team labels each text on a scale of 1-5 depending on the intensity of hate.

\begin{figure}[htbp]
    \hypertarget{RA3}{}
    \begin{tcolorbox}[colback=blue!7!white, colframe=blue!60!black, title=Research Agenda 3]
    Dataset creators must build on social science research and requirements of the task to define a schema that allows expressiveness and captures the complexity of hate.
    \end{tcolorbox}
\end{figure}
%in figure \hyperlink{RA1}{1.1}
%orange, red, purple, green, blue

\subsection{Defining the Annotation Guidelines}
\label{Defining the Annotation Guidelines}

Once the annotation schema is finalized it is the role of the annotation guidelines to provide instructions to the annotators on how to label each text. However, it is at this exact stage that human intervention in the process introduces subjectivity in the dataset. Any two annotators would have varying opinions on the same text owing to their backgrounds and life experiences. Depending on the requirement, it is imperative to define guidelines that specifically encourage or discourage this inherent subjectivity.

\citet{rottger-etal-2022-two} propose a framework of two contrasting data annotation paradigms differentiated on the basis of the intended downstream use of the dataset. They define the \textbf{descriptive} paradigm as one which encourages annotator subjectivity and allows the dataset to capture the diversity in the annotator's beliefs. On the other hand, they define the \textbf{prescriptive} paradigm to refer to modeling the annotation guidelines in a way that encodes one specific belief in the data and discourages annotator subjectivity. 

The authors state that explicitly sticking to one paradigm makes the future users of the annotated dataset clearly decide if it can or cannot be used for their task. 

To give a better idea of the different paradigms, let us take the example of task \textbf{T}. If the objective was to help design a content moderation system, then the authors would opt for the prescriptive paradigm when writing the guidelines. By referencing the hate speech laws in that region and the current political climate, a well-defined set of rules would give specific instructions to annotators when labeling each text. Further, the authors must provide gold-standard reference examples to help annotators learn the edge cases when labeling. However, in the research context of \textbf{T}, it is clear that authors want to value diversity in annotations. This means they don't need to design precise guidelines. Consider the prompt below:
\begin{quote}
    \textbf{P}: \textit{In your opinion, do you consider this text hateful towards sexual minorities?}
\end{quote}
Prompt \textbf{P} (coupled with a broad definition of \textit{hate}) could very well allow opposing annotations to seep into the dataset since different annotators would have different "opinions".

No paradigm is inherently superior. As long as the dataset creators correctly identify the downstream use of their dataset and define the guidelines according to any particular paradigm, it will be helpful for future users of that dataset to gauge if it can be useful for their work. For instance, if \textbf{D} was annotated based on a highly descriptive set of rules, it cannot possibly be used to train a model that is meant to only identify \textbf{\textit{explicit}} hateful comments against sexual minorities in a production setting since the dataset would contain instances of implicit hate as well.

It must be noted here that the choice of paradigm is strongly tied to the definition of hate that is formulated in Section \ref{Defining Hate}. For a descriptive paradigm, it might be preferable to define hate loosely using only criteria (1), (4), and (5) and a prompt \textbf{P} as specified above.

\begin{figure}[htbp]
    \hypertarget{RA4}{}
    \begin{tcolorbox}[colback=red!5!white, colframe=red!55!black, title=Research Agenda 4]
    Dataset creators must explicitly stick to either a descriptive or prescriptive annotation paradigm and release their annotation guidelines for future users to determine the feasibility of annotations for their use.
    \end{tcolorbox}
\end{figure}
%in figure \hyperlink{RA1}{1.1}
%orange, red, purple, green, blue

\subsection{Choosing Annotators with Specific Identities}

When recruiting annotators their socio-demographic traits must be taken into consideration. Often the individual identities and experiences of the annotators shape their decision on what they "perceive" as hate.

\citet{sap-etal-2022-annotators} found that black annotators were more likely to rate anti-Black posts as offensive than white annotators. Furthermore, they also showed a significant correlation between the annotators' conservative political leaning and their ratings of posts of African American English being racist.

\citet{Kim_Razi_Stringhini_Wisniewski_De_Choudhury_2021} studied the phenomenon of cyberbullying in online platforms. They analyzed the gap between outsider perspective (category label assigned by third-part annotators) and insider perspective (category label assigned by the post author) on the dataset and concluded that outsiders were more conservative than insiders and less likely to pick up on implicit references of bullying. They argued that labeling from the original victim could gather \textbf{more context-specific perspective} than that from random annotators who have been victims of bullying at some point of time, and definitely over third-party annotators with no prior cyberbullying experiences. 

Both these cases elicit the question:

\begin{quote}
    \textit{Must hate speech annotators have lived through the same life experiences or have the same sociodemographic background as the victims of hate?}
\end{quote}

The answer once again \textbf{depends on the task at hand} and the \textbf{allowed degree of subjectivity}. If it is needed to capture diverse views on the labels and the definition of hate is left to the annotators to some extent, it is beneficial to recruit annotators with "\textit{insider perspective}". However, if it is required to identify hate through a precisely formulated codebook, subjectivity is already reduced to a minimum and random third-party annotators can be employed. In this case, the strict guidelines would enforce high inter-annotator agreement in most instances which would otherwise be open to discussion. Thus it is acceptable to hire annotators who don't share similar life experiences.

We argue that in most cases the choice of determining which annotators to recruit roughly boils down to the choice of annotation paradigm (\S\ref{Defining the Annotation Guidelines}). A descriptive annotation paradigm could call for diversity in the annotator pool, or annotators sharing a specific set of identities with the target group. 

Thus one should take into account one or more of the following information to evaluate the impact of annotator differences \citep{vidgen2020directions, sap-etal-2022-annotators}:
\begin{enumerate}
\itemsep0em 
    \item Socio-demographic information
    \begin{itemize}
    \itemsep0em 
        \item Race
        \item Age
        \item Political Leaning
        \item Gender and/or Gender identity
        \item Sexual orientation
    \end{itemize}
    \item Professional expertise
    \item Personal experiences
    \begin{itemize}
    \itemsep0em 
        \item Experience of being a target of hate/abuse/bullying
        \item Experience observing expression of hate
        \item Experiences with hate in online settings
    \end{itemize}
\end{enumerate}

\begin{figure}[htbp]
    \hypertarget{RA5}{}
    \begin{tcolorbox}[colback=green!5!white, colframe=green!55!black, title=Research Agenda 5]
    Determining which annotators to hire is a decision that dataset authors should take into consideration and tie strongly with whether or not the added diversity be reflected or needed in the resulting dataset.
    \end{tcolorbox}
\end{figure}
%in figure \hyperlink{RA1}{1.1}
%orange, red, purple, green, blue

\subsection{Setting up the Annotation Process}

Ensuring that the labeling process is reliable and consistent is very important to the annotation process. With the increasing sizes of social media datasets, manual annotation has become cumbersome. Practitioners can use a number of commercially available platforms to aid the labeling process. 

Amazon mechanical Turk\footnote{\url{https://www.mturk.com/}} and Appen\footnote{\url{https://appen.com/}} (previously Figure Eight - F8) are popular choices for hiring crowdworkers \citep{caselli-etal-2020-feel, ousidhoum-etal-2019-multilingual, basile-etal-2019-semeval, Mollas_2022}. They provide granular control over filtering, training, and analyzing the performance of crowd workers to obtain desirable annotations.

When the dataset is large and annotators are not available, hiring crowd workers is a viable option. However, in some cases, researchers may employ non-expert annotators in the form of university students and graduate students or experts in the form of professionally trained annotators. In such a case it is possible to host a custom instance of the annotation platform for the annotators. doccano\footnote{\url{https://github.com/doccano/doccano}} \citep{doccano}, Potato\footnote{\url{https://github.com/davidjurgens/potato}} \citep{pei-etal-2022-potato}, Label Studio\footnote{\url{https://labelstud.io/}}, and Argilla\footnote{\url{https://argilla.io/}} are some easy to use open-source annotation platforms. They give you the option to configure the UI, onboard users, and aggregate labels. In collaboration with Huggingface spaces, Argilla allows you to host annotation instances for \textbf{free} based on limited compute. 

It must be noted that certain platforms as MTurk have geographical restrictions, thus it could be difficult to hire annotators specific to a particular region or from a specific culture. It is suggested to work with open-source platforms in that case instead.

\begin{figure}[htbp]
    \hypertarget{RA6}{}
    \begin{tcolorbox}[colback=purple!10!white, colframe=purple!70!black, title=Research Agenda 6]
    The choice of annotation platform is highly dependent on the size of the dataset and the availability of annotators.
    \end{tcolorbox}
\end{figure}

\subsection{Aggregating Labels}

\begin{table*}[t]
\begin{center}
\begin{adjustbox}{width=\textwidth}
\begin{tabular}{c c c c c c c} 
 \hline
 & \textbf{Define Hate?} & \textbf{Source} & \textbf{Schema} & \textbf{Paradigm} & \textbf{Annotators} & \textbf{Aggregation}\\ [0.5ex] 
 \hline
 \citet{toraman-etal-2022-large} & \ding{51} & Twitter & non-binary & H-P & non-expert & majority\\
 \hline
 \citet{grimminger-klinger-2021-hate} & \ding{51} & Twitter & binary & M-D & non-expert & N/A\\
 \hline
 \citet{caselli-etal-2020-feel} & \ding{51} & Twitter & non-binary & M-P & crowdsourced & majority\\
 \hline
 \citet{pamungkas-etal-2020-really} & \ding{51} & Twitter & binary & M-P & expert & majority\\
 \hline
 \citet{ousidhoum-etal-2019-multilingual} & \ding{51} & Twitter & multi-level & H-P & crowdsourced & majority\\
 \hline
 \citet{basile-etal-2019-semeval} & \ding{51} & Twitter & multi-level & H-P & crowdsourced & majority\\
 \hline
 \citet{zampieri-etal-2019-predicting} & \ding{51} & Twitter & multi-level & H-P & expert & majority\\
 \hline
 \citet{wijesiriwardene2020alone} & \ding{51} & Twitter & non-binary & H-P & semi-expert & majority\\
 \hline
 \citet{mathew2021hatexplain} & \ding{51} & Twitter + Gab & multi-level & H-P & crowdsourced & majority\\
 \hline
\citet{kennedy_atari_davani_yeh_omrani_kim_coombs_havaldar_portillo} & \ding{51} & Gab & multi-level & H-P & non-expert & majority\\
 \hline
 \citet{qian-etal-2019-benchmark} & \ding{51} & Reddit + Gab & multi-level & M-D & crowdsourced & majority\\
 \hline
 \citet{kurrek-etal-2020-towards} & \ding{51} & Pushshift Reddit corpus & multi-level & H-P & crowdsourced & majority\\
 \hline
 \citet{vidgen-etal-2021-introducing} & \ding{51} & Reddit & multi-level & M-P & expert + expert & majority\\
 \hline
 \citet{Mollas_2022} & \ding{51} & Reddit + YouTube & multi-level & M-P & crowdsourced & average\\
 \hline
 \citet{Sarkar_KhudaBukhsh_2021} & \ding{55} & YouTube & binary & H-D & non-expert & N/A\\
 \hline
 \citet{de-gibert-etal-2018-hate} & \ding{51} & Stormfront, & non-binary & H-P &  N/A & N/A\\
 \hline
 \citet{10.1145/3583562} & \ding{51} & Open source repositories & binary & H-P & non-expert & consensus\\
 \hline
 \citet{cercas-curry-etal-2021-convabuse} & \ding{51} & Chatbots & multi-level & M-P & semi-expert & individual\\
 \hline
 \citet{pavlopoulos-etal-2021-semeval} & \ding{55} & Civil Comments dataset & multi-level & M-D & expert & average\\
 \hline
 \citet{fanton-etal-2021-human} & \ding{51} & Initial seed + GPT2 & N/A & H-P & non-expert & N/A\\
 \hline
 \citet{hartvigsen-etal-2022-toxigen} & \ding{51} & GPT3 & binary & M-P & machine-generated & N/A\\
 \hline
 \citet{israeli-tsur-2022-free} & \ding{55} & Parler & Likert (1-5) & M-P & non-expert & average\\ [1ex] 
 \hline
\end{tabular}
\end{adjustbox}
\end{center}
\caption{22 publicly available hate speech datasets were analyzed for six of the 7 research agendas from Section \ref{Proposed Hate Speech Framework}. N/A indicates a lack of availability of sufficient information or the specific entry did not make sense. For \textbf{Paradigm}, the prefix H/M denotes \textit{highly/moderately} whereas the suffix P/D denotes \textit{Prescriptive/Descriptive} respectively.}
\label{table:1}
\end{table*}

The biggest challenge with building hate speech datasets is aggregating the labels after annotation. In fact, the question "\textit{how should we construct the ground truth from multiple labels?}" should be reframed to ask: 

\begin{quote}
    \textbf{Q}: \textit{whose truth is the ground truth and how do we distill it from multiple labels?}
\end{quote}
 
Since the perception of hate is subjective and varies from person to person, it depends on the stakeholders of the dataset as to what is the acceptable measure of ground truth, i.e., whose perspective should be considered as the gold standard. By stakeholders, we refer to the (1) the dataset creators (2) end users of the dataset, and (3) the annotators. To answer the latter part of \textbf{Q} we have to look at various approaches towards label aggregation. Considering the \textbf{majority} or \textbf{average} are the most common options. The annotation schema (\S\ref{Defining the annotation schema}) generally limits how labels can be aggregated. 

Let us consider a categorical binary/non-binary taxonomy of hate for task \textbf{T}. In such a case it is very difficult to aggregate labels through averaging and the majority vote is generally considered acceptable. It then depends if taking the majority vote for each text serve the stakeholders well. For a highly prescriptive downstream use case, most annotators would be required to have similar views, low inter-annotator agreement would be a red flag, and majority voting is desirable.

Ordinal labels, on the other hand, give greater flexibility in terms of conversion to numeric scales and averaging of labels. A highly descriptive downstream use case (such as a critical analysis of the annotator's response to controversial BLM tweets) would benefit from such aggregation. 

Minority labels are important carriers of opposing views and, in the case of a diverse annotator pool, should not be overridden by majority consensus. \citet{Mitchell_jury} introduce \textit{jury learning} as a way to \textbf{model each annotator in the dataset}. By sampling a set of jurors based on their characteristics, a model trained using jury learning then returns the median jury outcome over \textit{N} trials. The architecture of jury learning affords practitioners to contextualize predictions in social variables and allows interpretability in decision-making. \citet{davani-etal-2022-dealing} propose a similar multi-annotator modeling architecture based on a multi-task approach in which they predict individual labels corresponding with each annotator for each text.

\begin{figure}[htbp]
    \hypertarget{RA7}{}
    \begin{tcolorbox}[colback=yellow!10!white, colframe=yellow!70!black, title=Research Agenda 7]
    The ground truth should reflect the perception of the stakeholders and the choice of label aggregation technique depends on the annotation schema. A good aggregation approach values annotator diversity and minority opinion unless the guidelines expect annotators to agree substantially.
    \end{tcolorbox}
\end{figure}
%in figure \hyperlink{RA1}{1.1}
%orange, red, purple, green, blue

\section{Challenges}
\label{Challenges}

Table \ref{table:1} summarizes some of the popular hate speech datasets taken from \url{hatespeechdata.com/} (except \citet{israeli-tsur-2022-free}) by grounding them across six of the seven agendas we discussed in Section \ref{Proposed Hate Speech Framework}. In continuation with the discussion in previous sections, we identify three main challenges in today's work on hate speech detection:
\begin{enumerate}
\itemsep0em 
    \item \textbf{Subjectivity and Annotator Bias.} Subjectivity in annotations is inevitable and cannot be eliminated even through a well-constructed definition of \textit{hate} and deft guidelines. Similarly, it is impossible to completely remove the effect of annotators' personal background from reflecting in their work, although skillful training, providing borderline examples, and hiring experts can help reduce the extent of bias and subjectivity.
    \item \textbf{Data Statement.} To ensure that datasets can be used fairly and reliably in the future well-defined data statements must be incorporated into metadata already associated with data sets \citep{bender-friedman-2018-data}.
    \item \textbf{Context.} To further reduce ambiguity and inconsistency in annotations, practitioners should try to provide the context along with independent text samples.
\end{enumerate}

\section{Conclusion}
\label{Conclusion}

In this paper, we discussed the challenges surrounding hate speech detection through the point of view of data. Based on the different stages encountered when creating a dataset, we identified seven points of dissent that call for making sound decisions based on the task at hand and the interests of the stakeholders. We proposed a framework outlining distinct research agendas for each of these seven points and made arguments supporting the different choices that can change the polarity and consistency of the dataset. Finally, we asked dataset creators to embrace subjectivity in annotations, release data statements with their datasets, and try and give context to their datasets for each data point. We hope that this framework will provide researchers with the best practices and equip them to ask pertinent questions when building a dataset for hate speech detection in the future.

\section*{Limitations}
Although we tried to present a thorough data-centric approach to analyzing the process of dataset creation, we did not discuss certain design options as part of our framework. For example, when sourcing social media data, dataset sampling is almost always performed to offset the inherent class imbalance. Such data engineering techniques were out of the scope of our discussion and call for a separate granular work. Thus our research agenda is not exhaustive.

In addition, we have only taken textual data into consideration. However, data in the wild is not uni-modal. We do hypothesize that our framework can be easily generalized to other modalities although certain modalities such as images could elicit additional research agendas while building the dataset. 

Finally, we have only focused on English datasets for our analysis. A more rigorous study encompassing multilingual hate speech needs to be conducted to include other languages.

\section*{Ethics Statement}
This work focuses on current challenges in building hate speech datasets. Our proposed
research agendas are specific to the literature in this domain and complement existing frameworks, to encourage future researchers to think carefully about the design choices to consider and questions to ask.

All datasets used in this study are publicly available and used under strict ethical guidelines. We only use attributes that the authors have self-identified in our experiments. We ask dataset creators to use the framework in this work to increase the fairness and reliability of the annotations in their dataset and propagate ethical and useful work of their data.

% \section*{Acknowledgements}
% This document has been adapted by Jordan Boyd-Graber, Naoaki Okazaki, and Anna Rogers from the style files used for earlier ACL, EMNLP, and NAACL proceedings, including those for

\bibliography{anthology,custom}
\bibliographystyle{acl_natbib}

\end{document}